\documentclass{article}
\usepackage[T1]{fontenc}
\usepackage[utf8]{inputenc}
\usepackage{graphicx}
\usepackage{subfigure}
\usepackage{caption}
\usepackage[dvipsnames]{xcolor}
\usepackage{geometry}
\usepackage{authblk}
\usepackage{pdfpages}
\usepackage{hyperref}
\title{Surgical Visual Domain Adaptation: Results from the MICCAI 2020 SurgVisDom Challenge}
\author[1]{Aneeq Zia}
\author[1]{Kiran Bhattacharyya}
\author[1]{Xi Liu}
\author[1]{Ziheng Wang}
\author[2]{Satoshi Kondo}
\author[3]{Emanuele Colleoni}
\author[3]{Beatrice van Amsterdam}
\author[4]{Razeen Hussain}
\author[5]{Raabid Hussain}
\author[6]{Lena Maier-Hein}
\author[3]{Danail Stoyanov}
\author[7]{Stefanie Speidel}
\author[1]{Anthony Jarc}
\affil[1]{Intuitive, Inc.}
\affil[2]{Konica Minolta, Inc.}
\affil[3]{University College London}
\affil[4]{University of Genoa}
\affil[5]{University of Burgundy}
\affil[6]{German Cancer Research Center (DKFZ)}
\affil[7]{National Center for Tumor Diseases (NCT)}

\date{}

\begin{document}

\maketitle

\begin{abstract}
Surgical data science is revolutionizing minimally invasive surgery by enabling context-aware applications. However,  many challenges exist around surgical data (and health data, more generally) needed to develop context-aware models. This work - presented as part of the Endoscopic Vision (EndoVis) challenge at the Medical Image Computing and Computer Assisted Intervention (MICCAI) 2020 conference - seeks to explore the potential for visual domain adaptation in surgery to overcome data privacy concerns. In particular, we propose to use video from virtual reality (VR) simulations of surgical exercises in robotic-assisted surgery to develop algorithms to recognize tasks in a clinical-like setting. We present the performance of the different approaches to solve visual domain adaptation developed by challenge participants. Our analysis shows that the presented models were unable to learn meaningful motion based features form VR data alone, but did significantly better when small amount of clinical-like data was also made available. Based on these results, we discuss promising methods and further work to address the problem of visual domain adaptation in surgical data science. We also release the challenge dataset publicly at \url{https://www.synapse.org/surgvisdom2020}.
\end{abstract}

\section{Introduction} 
The goal of surgical data science is to improve the quality and efficacy of surgical care through the collection, analysis, and modeling of data \cite{maier2017surgical,vedula2017surgical, maier2020surgical}. Some of the key data modalities are images and videos \cite{maier2017surgical, esteva2021deep}. Consequently, within the medical imaging and computer assisted interventions domain, there have been multiple challenges organized to target problems in machine vision \cite{endovis_2020,lesion_seg_2018,retina_2020,rib_2020,maier2020bias, allan2017, allan2018}. Specifically, surgical activity detection and workflow analysis have a history of approaches and recent interest \cite{padoy2019machine,morita2019real,khalid2020evaluation,kitaguchi2020real,tanwani2020motion2vec,sharghi2020automatic, zia2017temporal, zia2018surgical}.

Being able to automatically recognize different steps or tasks within a surgical procedure can lead to development of many exciting context-aware applications, like task-based performance reports, OR management, and  real-time surgeon augmentation \cite{padoy2019machine,brown2020bring,dias2021augmented, zia2019novel}. However, the lack of publicly available large-scale labelled surgical datasets for the community remains an issue \cite{maier2017surgical, maier2020surgical}. One of the main reasons for this lack of data is the sensitivities around surgical data and personal health information which makes it hard to collect and share \cite{gostin2018health}. Other reasons like difficulties in surgical data acquisition and annotation also cause problems in sharing such data \cite{maier2020surgical}. 

To address this issue, we proposed a challenge to see whether virtual reality (VR) simulations of robotic-assisted surgical tasks can be leveraged to learn transferable representations of surgical tasks for real-life settings. In the broader computer vision community, this research problem is framed as visual domain adaptation \cite{wang2018deep, peng2018visda} and already has some preliminary implementations in the medical domain for image classification and segmentation \cite{ross2018exploiting, mahmood2018unsupervised, sahu2020endo}. Specifically, our aim was to see if relevant motion features can be learned from videos of VR surgical tasks and then be used to recognize those surgical tasks in clinical-like settings. Our analysis of participants models showed that supplementing a small proportion of clinical-like data with VR data significantly helps the models to learn generalizable features. We discuss in detail the performance of each method in recognizing different surgical tasks and propose future work to further the field of visual domain adaptation in surgery.

\section{Surgical Visual Domain Adaptation Challenge}
\subsection{Overview}
This challenge was organized as a sub-challenge \footnote{https://www.synapse.org/surgvisdom2020} under the Endoscopic Vision Challenge \footnote{https://endovis.grand-challenge.org/} at MICCAI 2020 \footnote{https://miccai2020.org/}. The design of the challenge was based on BIAS standards \cite{maier2020bias} (the full design document can be found in Appendix A). We divided the challenge into two categories. The first category required the participants to use a mix of VR videos and a small number of clinical-like videos to train their models and make predictions on clinical-like videos (soft-domain adaptation). Whereas, for the second category, the participants were required to use only VR data when training their models and were tested on the same clinical-like videos as before (hard-domain adaptation). 

\begin{figure}[t]
\begin{center}
  \includegraphics[width=0.8\linewidth]{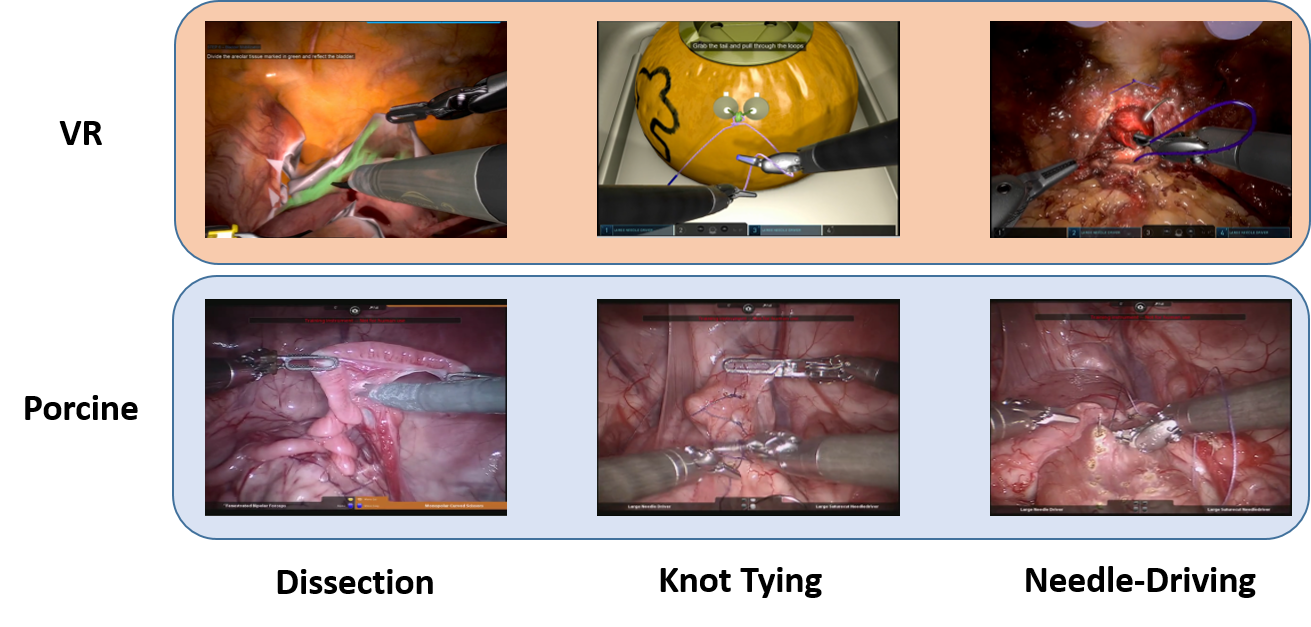}
  \caption{Sample frames from the dataset for each surgical task within VR and porcine domains.}
  \label{fig:dataset_sample_images}
\end{center}
\end{figure}

\begin{figure}[t]
\begin{center}
  \includegraphics[width=0.8\linewidth]{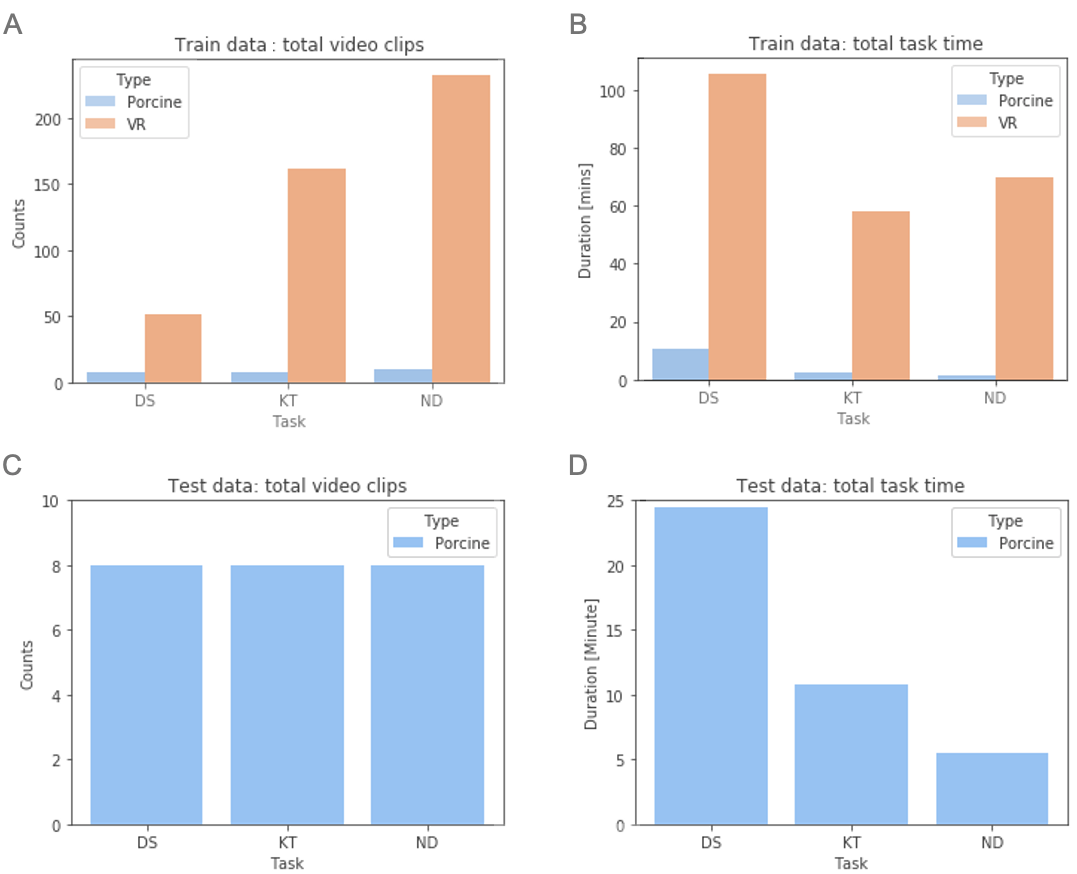}
  \caption{Training and testing data set characteristics used for both categories in the challenge. A) The total number of video clips per task in VR and clinical-like porcine domains for the training set. B) The total duration of video per task in VR and clinical-like porcine domain for the training set. C) Total number of videos in the testing set from clinical-like porcine domain where each task appears. D) Total duration of task time in the testing set in the porcine domain.}
  \label{fig:dataset_summary}
\end{center}
\end{figure}

\subsection{Data Description}
The dataset consists of videos from two domains: virtual reality (VR) and clinical-like (surgical tasks performed on a porcine model). All videos were captured from either a da Vinci robotic system (Xi or Si) or the da Vinci simulator.There are a total of 3 surgical tasks from both VR and a porcine model: dissection (DS), knot-tying (KT), and needle-driving (ND). The subjects performing these tasks ranged from beginners with no surgical expertise to expert surgeons. Figure \ref{fig:dataset_sample_images} shows some sample frames from the dataset.

The training dataset consists of video clips where subjects performed any one of the 3 tasks. Each video clip contains only one task with varying duration, i.e. each training video clip has a single task label (ND, KT, DS). In the training dataset, video clips extracted from both VR skill exercises and a porcine model were provided. The VR exercises were completed using a da Vinci simulator. In addition to VR videos, a small set of video clips of analogous surgical tasks performed on a porcine model are provided. The VR videos were captured at 60 fps with a resolution of 720p ($1280\times720$) from one channel of the endoscope. The porcine videos were captured at 20 fps with a resolution of $960\times540$ from one channel of the endoscope. In total, the training set contains 450 clips from VR exercises and 26 clips from a porcine model as detailed in Figure \ref{fig:dataset_summary}.

The testing dataset consists of 16 video clips from the porcine model only, where the subject performed at least one of the tasks. Eight video clips in the test set contained more than one surgical task with no overlap – two tasks did not occur at the same time. Specifically, KT and ND tasks co-occurred in eight video clips, and DS tasks occurred in the other eight videos. Any periods of inactive time in the video or other tasks that did not qualify as one of the 3 surgical tasks used here were not used to evaluate challenge submissions. The challenge participants had to produce frame-by-frame predictions for each video clip in the test set.

This dataset is now released publicly and can be downloaded from \url{https://www.synapse.org/surgvisdom2020} and used for non-commercial purposes.

\subsection{Challenge Categories}
Two challenge categories are proposed to evaluate participating team's model. For each category, teams were not allowed to use any other surgical dataset, public or private, at any step of their training process. However, they were allowed to use publicly available vision datasets for pre-training models, like ImageNet \cite{krizhevsky2012imagenet} and Kinetics \cite{kay2017kinetics}.

\subsubsection{Category 1: Soft domain-adaptation}
Challenge participants were asked to use the entire training dataset--both VR and porcine videos--to train their machine learning models to recognize the 3 surgical tasks. As there are some porcine videos in the training set ($\sim$10\%), we call this the ``soft domain-adaptation'' problem. The performance of these models was then evaluated on the test dataset with videos of only surgical tasks on porcine model. 

\subsubsection{Category 2: Hard domain-adaptation}
Challenge participants were asked to train their models on only the VR video clips in the training dataset and exclude the porcine video clips. Since there are no porcine video clips in the training set, we call this the ``hard domain-adaptation'' problem. The performance of these models trained only of VR data was then evaluated on the test dataset with only porcine videos.

\section{Participating Teams Methods}

\subsection{Team Parakeet (University of Genoa and University of Burgundy)}

\begin{figure}[t]
\begin{center}
  \includegraphics[width=0.8\linewidth]{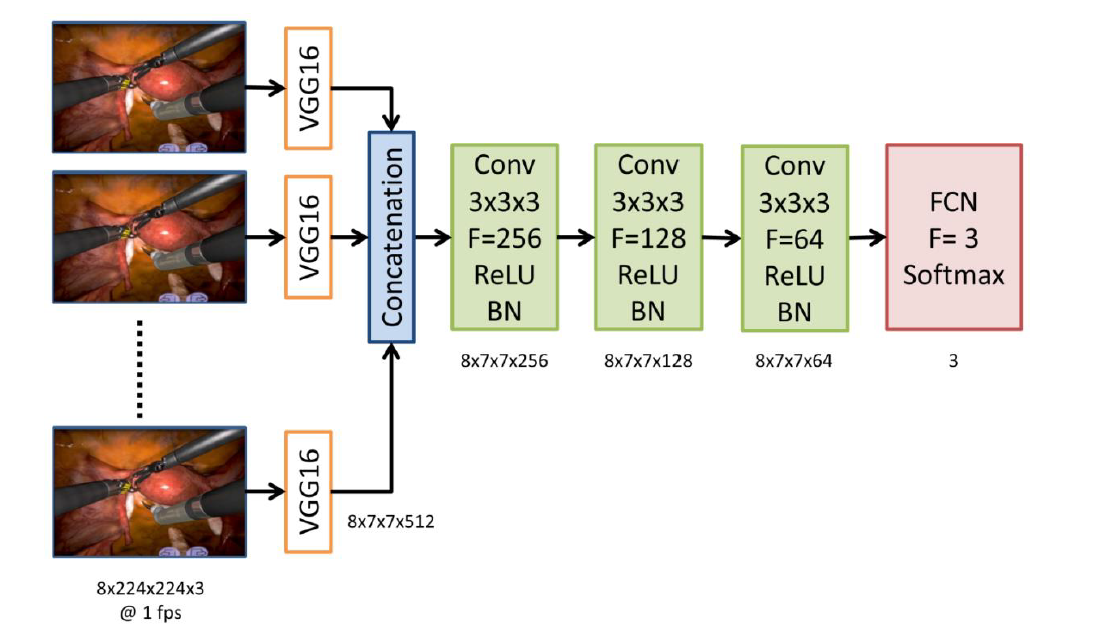}
  \caption{Team Parakeet's proposed spatio-temporal model architecture. VGG16: 2D VGG16 pretrained on imagenet dataset, Conv: 3D convolutional layer, BN: batch normalization, FCN: fully connected layer, F: number of features.}
  \label{fig:model_parakeet}
\end{center}
\end{figure}

Team Parakeet performed some pre-processing of data before feeding it into their model training. Since the fps of the VR and porcine videos was significantly different, the training videos were downsampled to 5 fps. This frame rate was chosen as the videos consisted of moderate movements which do not require a high refresh rate for determining movements. For memory limitations, the frames were also desampled to a lower resolution of 224x224x3 pixels. The video frames were input to a 2D-3D convolutional neural net (CNN) depicted in Figure \ref{fig:model_parakeet}. This team's proposed architecture takes sequence of 8 frames as input (8x224x224x3) and outputs the softmax classification scores. First 2D VGG16 features (7x7x512) were extracted for each video frame individually. The VGG network was pretrained using the imagenet dataset \cite{krizhevsky2012imagenet}. The VGG16 features for the sequence were passed through a 3DConvNet to integrate spatial and temporal information. The 3DConvNet consisted of 3x3 convolutional layers followed by a fully connected layer and was trained from scratch. All convolutional layers had ReLU activations followed by batch normalization. Categorical-crossentropy was used as the loss function. A 40\% dropout was used before the final fully connected softmax layer. The filter sizes are shown in Figure \ref{fig:model_parakeet}.
The architecture was implemented on a computer with dedicated GPU (NVIDIA GeForce GTX 1080, 8 GB RAM processor) using Keras and Tensorflow libraries. For category 2, only a training dataset (without validation) was used containing all the VR video frames. Whereas for category 1, some porcine model videos were included in the training dataset along with VR videos with a training-testing split ratio of 60:40. The training was performed for 500 epochs with a batch size of 32 using Adam optimizer with a learning rate of 0.0001.
During training, a data augmentation strategy was adopted in which the video stream was divided into fixed size segments. The input frames for the proposed VGG-3DConvNet were randomly selected from within each segment. Thus, the input to the network was at 1 fps with data augmentation performed on 5 fps data stream.

\subsection{Team SK (Konica Minolta, Inc.)}
\begin{figure}[t]
\begin{center}
  \includegraphics[width=0.8\linewidth]{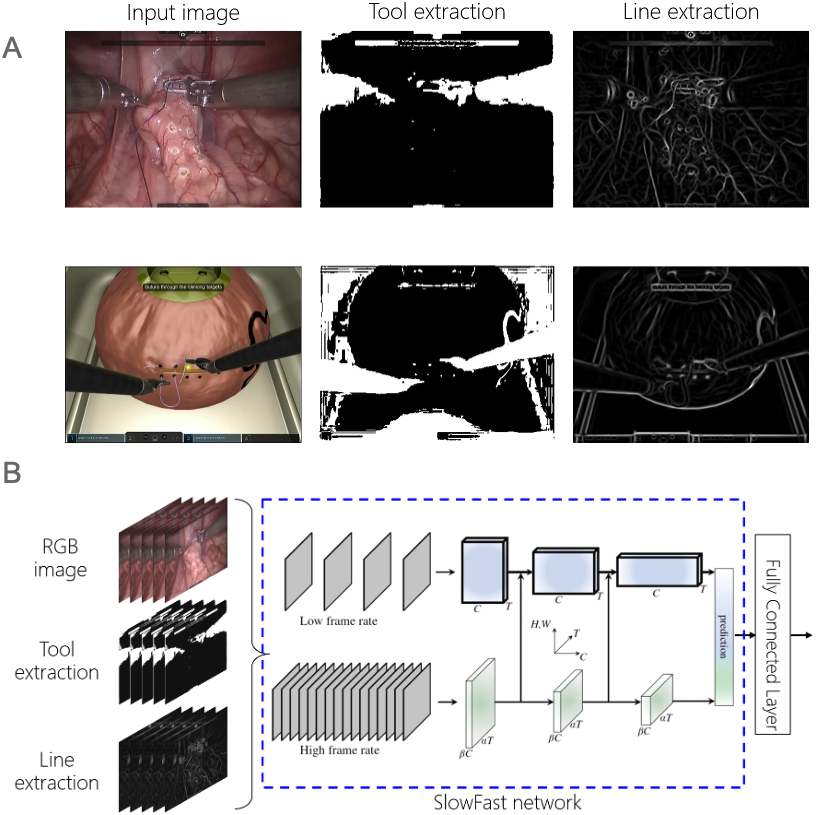}
  \caption{A) Pre-processing steps for team SK. Color-variance based segmentation of tools (top) and line filter based segmentation for surgical needle (bottom). B) Proposed model for team SK. SlowFast network (taken from \cite{feichtenhofer2019slowfast}) pre-trained on Kinetics 400.}
  \label{fig:combo_SK}
\end{center}
\end{figure}

The second participating team consisted of Satoshi Kondo from Konika Minolta Inc, Japan. The method proposed by them was based on 3D deep neural network. In order to utilize the image features of surgical tools and suture needle, two types of pre-processing were implemented as shown in Fig \ref{fig:combo_SK}A. The first type of pre-processing involved detection of surgical tools based on image processing. The main colors of the surgical tools are silver and black where the values in RGB for those colors are very similar. Therefore, the standard deviation of RGB values for each pixel was calculated and areas where the standard deviation was below a certain threshold were masked. This threshold was set as 10 (as the pixel value is expressed in 8 bits). The second type of pre-processing used a line filter \cite{sato19973d} to extract suture needle. With preprocessing, there are three cases in terms of number input channels. In the first case, the original image by itself has three input channels. In the second case, the original image is concatenated with one of the pre-processed images resulting in four input channels. In the final case, the original image  is used along with two pre-processed images resulting in five input channels.

For the proposed model, SlowFast network \cite{feichtenhofer2019slowfast} was used as a base network as shown in Fig \ref{fig:combo_SK}B. The base network was pre-trained with Kinetics-400 dataset \cite{kay2017kinetics} and a fully connected layer was added at the end to output three classes. In order to cater for the data imbalance, over-sampling and under-sampling were performed on porcine and VR training videos, respectively.
The training method used by this team is as follows. An input image is resized to 320 x 240 after the border area is cropped. The frame rate is sub-sampled to 5 fps and the number of frames for the 3D deep neural network is 32. Lookahead optimizer \cite{zhang2019lookahead} is used in training with an initial learning rate of 1e-7 for Category 1 and 1e-8 for Category 2. The learning rate is updated with cosine annealing at each epoch.
The mini-batch size is kept as 12 and the experiments are run for 30 epochs using softmax cross entropy loss. Data augmentation like translation, rotation, resizing and contrast adaptations, are also applied on the fly during the training. The final model selected after experimentation used five channel inputs for Category 1 and four channel inputs, the original image and the mask image, for Category 2. At inference, a moving average filter was used with a length of 4 seconds.

% Ziheng: edit script, summarize methods, change the first person 'we' to 'the team', check typo, add reference 
\subsection{Team ECBA (University College London)}
\begin{figure}[t]
  \includegraphics[width=\linewidth]{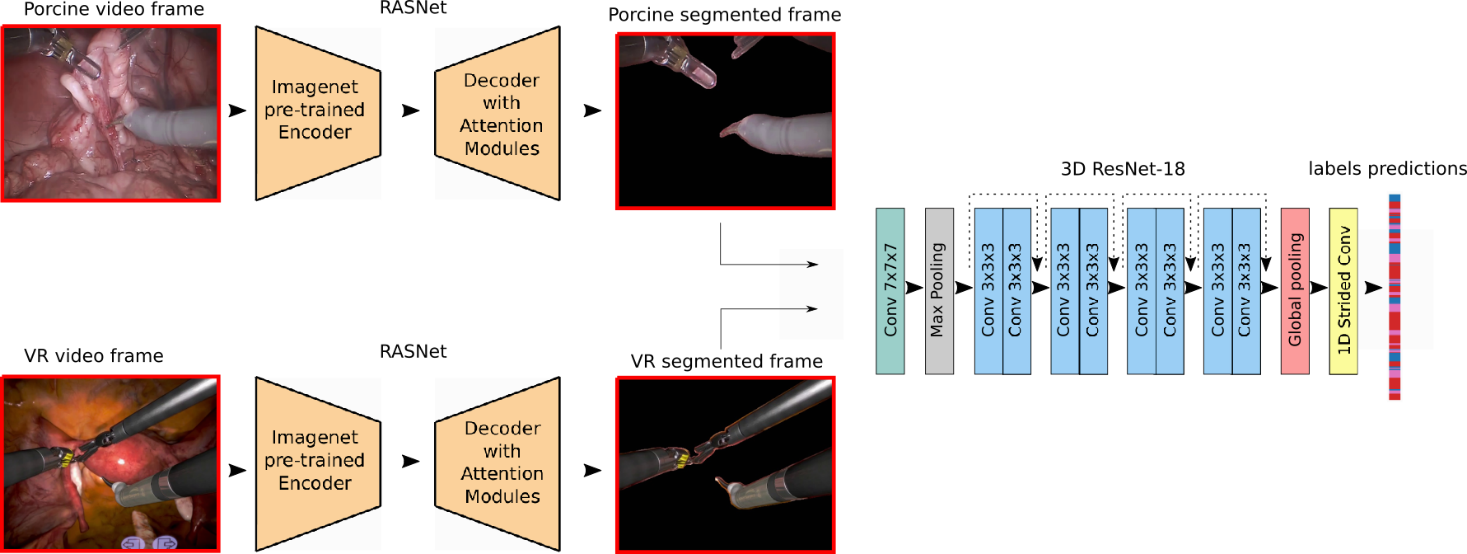}
  \caption{Workflow of the presented method of Team ECBA. First two segmentation models (RASNet~\cite{ni2019rasnet}) were trained on manually segmented images (one on virtual reality and the other on porcine frames respectively). Then, once all the video images were segmented, a 3D ResNet-18 for action recognition  was trained to predict on test frames.}
  \label{fig:team_ECBA}
\end{figure}

Team ECBA hypothesized that the domain gap between real and VR images can be reduced by removing the background of endoscopic views, which was claimed to be the one of the main differences between the two domains. They proposed to address the domain adaption for surgical action recognition in two separated steps, as shown in Figure~\ref{fig:team_ECBA}. First, the team removed the background pixels from both VR and porcine images using a segmentation network trained on manually segmented frames. Then, a 3D action recognition network was trained on the segmented frames to predict surgical actions on the given test set. Finally, once the predictions on the test videos were obtained, the team applied a classification filter on them to avoid rapid class bouncing due to the action changes inside a single video. 

\subsubsection{Segmentation network}
Since no segmentation ground truth was provided for the videos in this challenge, the team manually segmented 300 frames for each category (VR and Porcine) using VGG image annotator tool. Inside each category, the team selected the segmentation frames to be balanced in terms of performed action, i.e. 100 frames were selected for each action (Dissection, Needle driving and Knot Tying). For each video frame, we segmented both tools and needles, leaving all the rest as background.
Before manual segmentation, each frame was first cropped to remove camera side artifacts and then resized to 384x480. Images were finally randomly split into training (270 frames) and validation set (30 frames).
The team chose RASNet~\cite{ni2019rasnet} as the segmentation models (one for VR and one for porcine frames). This architecture has the famous U-net~\cite{ronneberger2015u} structure and uses an Imagenet pre-trained ResNet~\cite{he2016deep} as feature extractor. Moreover, in the decoding part, several refined attention blocks are employed along with classic convolutional layers and skip connections to allow the network to properly focus on key regions in the image.
Since the dataset the team used to train the segmentation networks did count only 300 labelled frames for each category, they augmented the data by randomly flipping them both horizontally and vertically. The team trained their model using the sum of binary cross entropy and Intersection over Union (IoU) score as our loss function, that has been shown to be particularly effective for tool segmentation~\cite{ni2019rasnet, iglovikov2018ternausnet}. The team chose Adam~\cite{kingma2014adam} as optimizer with learning rate $0.001$, a batch size of 4, and trained both models for 300 epochs, selecting the best as the one that maximized the IoU score on validation data.

\subsubsection{3D action recognition network}
Once both VR and porcine segmentation models were trained, the team trained the 3D action recognition network on the segmented frames. For this task, the team selected the architecture proposed in ~\cite{funke2019using}: here, the authors proposed a 3D ResNet-18 trained on Kinetics dataset for surgical gesture recognition. Although in their work Funke I. et al chose a temporal window of 16 frames recorded at 5 frames per second (fps), the team doubled this window to 32 to increase the the temporal field of view of the network, thus allowing it to process 6 second of video in one shot. The team trained the action recognition network using multi-class cross entropy loss with Adam optimizer, a batch size of 32 and a learning rate 0.001.
In order to train the network, the team split the segmented frames into train and validation sets. Since validation data should be a representative sample of the test set, the team chose two porcine videos from each action folder as validation videos, leaving all the remaining data (remaining porcine videos and all VR videos) for training. Again, the team selected the model that obtained the best loss scores on validation data.
Finally, once the predictions on the test set were obtained, a classification filter was applied to avoid rapid class bouncing, that is particularly present during action changes inside a single video. For each video frame the filter applies a temporal window of length W that covers $-W/2$ past frames and $W/2$ future frames. The new class is selected as: $new Ci = maxCount(C_{i+t})$ for $t \in[-W/2, W/2]$, where $i$ is the selected frame, $C$ is the class predicted by the 3D network and $new C$ is the class after filtering. In words, the filter selects the most frequent class inside the window and labels the central time frame with that class. Since surgical actions usually last from 5 second on, the team selected a temporal window with the size $W=100$. Considering that the test videos were recorded at 20 frames per second, the selected temporal window is 5 seconds large.

\section{Results}

\subsection{Evaluation Criteria}
For each video clip in the testing set, we computed the weighted average f1-score \cite{scikit-learn} for the predicted class labels as provided by the challenge participants. The f1-score is computed in the following way for each label in the test video clip,

\begin{equation}
    f1 = 2 * \frac{precision * recall}{precision + recall}
    \label{eq:f1_score}
\end{equation}

Then the f1-score for each label present in the video was averaged weighted by support (the number of true instances for each label). Teams were ranked by this mean f1-score from all videos to determine the winner and runner-up for each category. However, to test rank stability, we also computed unweighted average f1-scores, global f1-scores, and balanced accuracy scores for each test video clip and, subsequently, ranked teams by their mean for these metrics across all test videos. We found no dramatic differences in team ranking when using these other metrics. 

For the sake of comparison, we have also included predictions from a random model that was equally likely to predict any of the 3 labels for each frame of video. The predictions from the random model serve as a baseline for the precision and recall performance of challenge submissions. In all figures below, we refer to random model as `Rand'. 

\subsection{Category 1}
\begin{figure}[t]
  \includegraphics[width=\linewidth]{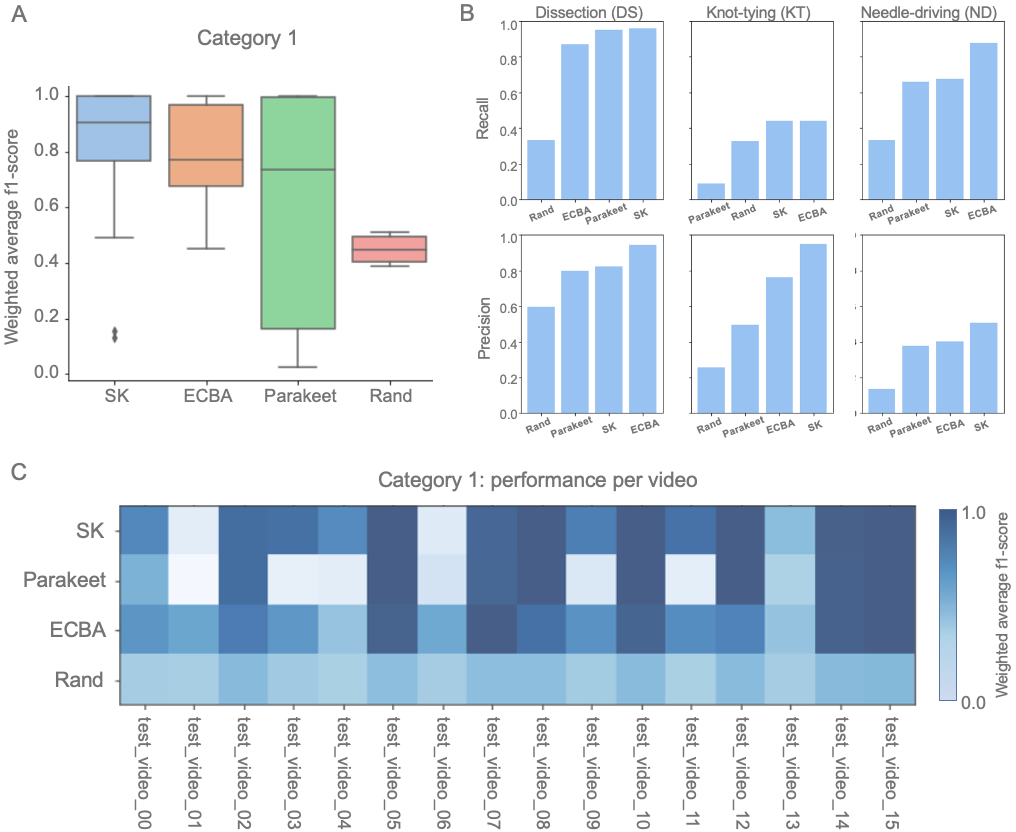}
  \caption{Category 1 Results. A) Distribution of weighted average f1-scores of each team for all videos in the test set. B) Average recall and precision for each team on each task. C) Weighted-average f1-score for each team for each video.}
  \label{fig:cat_1_box}
\end{figure}

For Category 1, where the training set included some porcine videos, all teams outperformed the random model (Rand), as can be seen in the distributions of weighted average f1-scores for each team in Figure \ref{fig:cat_1_box}A. Team SK outperformed ECBA and both teams outperformed Parakeet. However, Parakeet had a large variance in weighted average f1-scores that sometimes under-performed the random predictions.

In Figure \ref{fig:cat_1_box}B, we can see the precision and recall of each team broken down by task labels averaged across all videos. All submissions had similarly high precision and recall for the Dissection task when compared to the random predictions. However, the submissions tended to have higher precision than recall for the Knot-tying task and higher recall than precision for the Needle driving task. In nearly all cases, submissions outperformed the precision and recall of the random predictions. Nevertheless, the recall of the submissions for the Knot-tying label was closest to the baseline provided by the random predictions, suggesting that this was the most challenging metric for the submissions. 

Upon closer inspection of the models' performances on each test video clip in Figure \ref{fig:cat_1_box}C, we can see that some test video clips were more difficult than others for all of the model submissions. For instance, all submissions performed only slightly better than `Rand' for test\_video\_clip\_0013. Similarly, test\_video\_clip\_0001 and 0006 also had drops in model performance across all submissions. These videos where the model performances dropped included the Needle-driving and Knot-tying tasks but not Dissection, mirroring the task-based results seen in Figure \ref{fig:cat_1_box}B. 

On the other hand, there were some test videos where all the teams showed strong performance far above the `Rand' predictions. For instance, for test\_video 0010, 0014, and 0015, all submissions had an average weighted f1-score of nearly 1.0 suggested that most frames were predicted correctly. The videos where the models perform dramatically better than random were videos with the Dissection task label.

When ranking the teams by overall performance we found that the mean value of various different metrics produced the same ranking as seen in table with Category 1 rankings. All submissions consistently outperformed the `Rand' predictions by large margins for all metrics we computed. Team SK consistently outperformed ECBA but, for some metrics, only by a small margin. For example, the mean balanced accuracy for Team SK and ECBA were nearly the same value. On the other hand, the unweighted f1-score emphasized the difference between each team the most and also had the lowest value for the random prediction model. 

\begin{center}
\begin{tabular}{ |c|c|c|c|c| }
 \hline
 \multicolumn{5}{|c|}{Category 1 rankings (average for test set)} \\
 \hline
 Team & weighted f1-score & unweighted f1-score & global f1-score & balanced accuracy \\ 
 \hline\hline
 Rand & 0.45 & 0.207 & 0.327 & 0.327 \\  
 \hline
 Parakeet & 0.60 & 0.414 & 0.599 & 0.644 \\
 \hline
 ECBA & 0.79 & 0.488 & 0.742 & 0.776 \\
 \hline
 SK & 0.80 & 0.604 & 0.774 & 0.778 \\
 \hline
\end{tabular}
\end{center}

\subsection{Category 2}
\begin{figure}[h]
  \includegraphics[width=\linewidth]{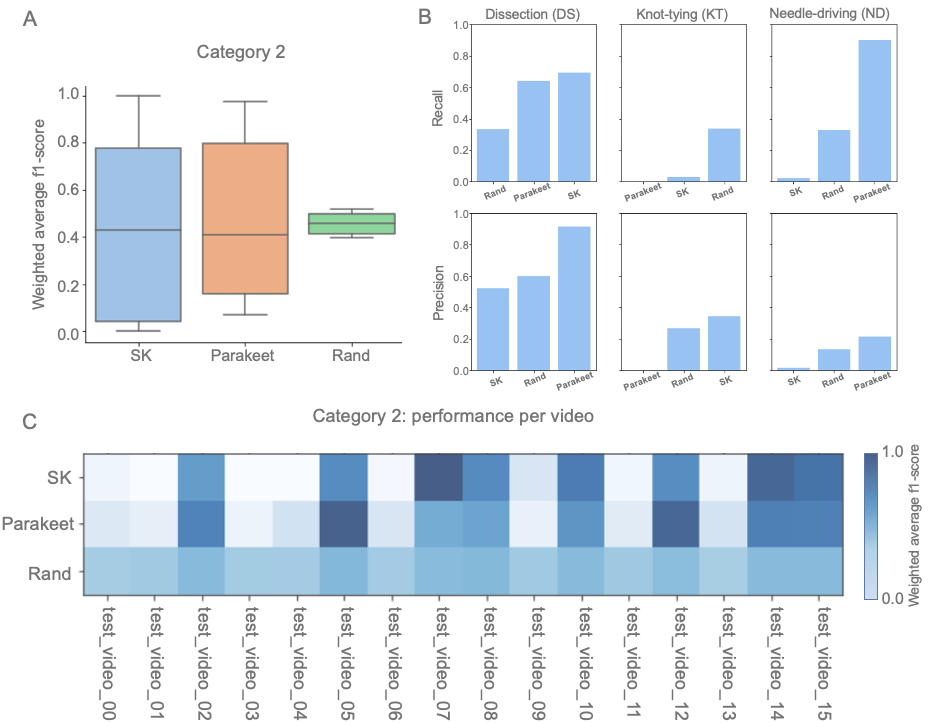}
  \caption{Category 2 Results. A) Distribution of weighted average f1-scores of each team for all videos in the test set. B) Average recall and precision for each team on each task. C) Weighted-average f1-score for each team for each video.}
  \label{fig:cat_2_box}
\end{figure}

For category 2, where the training set only had virtual reality videos, the submissions had large variances in their performance as seen in Figure \ref{fig:cat_2_box}A. Team SK and Parakeet had similar distributions of weighted average f1-scores across all the videos and much more variance than random predictions. 

When inspecting the precision and recall of each model grouped by the surgical task label in Figure \ref{fig:cat_2_box}B, we find that the submissions don't consistently outperform the random predictions. However, the precision and recall of the submission from team Parakeet outperforms the random predictions for the Dissection and Needle-driving tasks.

The mixed results in precision and recall are likely because submissions tended to perform far better on some test video clips than others as seen in Figure \ref{fig:cat_2_box}C. The subset of test video clips for which the submissions outperformed random predictions included only the Dissection task label. However, for the test videos which included the Knot-tying and Needle driving tasks, the submissions often under-performed the random predictions, suggesting the occurrence of over-fitting during model training. 

These differences between random predictions and submissions are further summarized in the Category 2 rankings table. The mean value of various different metrics produced the same rankings with submissions outperforming the random predictions overall with different margins. For instance, the value of the mean weighted f1-scores only had small differences between the teams. However, balanced accuracy and global f1-score showed a much larger gap between team Parakeet and SK.

\begin{center}
\begin{tabular}{ |c|c|c|c|c| }
 \hline
 \multicolumn{5}{|c|}{Category 2 rankings (average for test set)} \\
 \hline
 Team & weighted f1-score & unweighted f1-score & global f1-score & balanced accuracy \\ 
 \hline\hline
 Rand & 0.45 & 0.207 & 0.327 & 0.327 \\  
 \hline
 SK & 0.46 & 0.225 & 0.370 & 0.369 \\
 \hline
 Parakeet & 0.47 & 0.266 & 0.475 & 0.559 \\
 \hline
\end{tabular}
\end{center}

\section{Discussion}

Overall, the teams performed better on the dissection tasks in both categories. For category 1, all teams performed better than random suggesting that all models were able to learn generalizable patterns in the dataset. More specifically, in Category 1, Team SK and Team ECBA performed similarly in metrics but actually had different strengths as seen by their different precision and recalls in Figure \ref{fig:cat_1_box}B. This means that a fusion of these two models can potentially have some added benefit and perform better than the individual models. 

For Category 2, the teams performed worse than they did in Category 1, but still better overall than random predictions suggesting that there were still some generalizable patterns learned. Team Parakeet showed stronger performance in Dissection and Needle-driving suggesting that the VR domain provided similar features to porcine domain for these two tasks. 

Comparing different methodologies, it seems like the approach taken by Team ECBA and Team SK of using tool detection and having networks pretrained on Kinetics dataset, helped them in Category 1, overall. However, these techniques did not lead to better performances for Category 2. For team Parakeet, using VGG network to extract features before learning temporal features seemed to help them perform reasonably well for dissection and needle-driving in both categories. This suggests that a mixed method of combining multiple model architectures may be needed to discern different surgical tasks from different modalities.

It is important to note that the overall lack of performances could potentially be due to the relatively small size and the class distribution of the dataset. The per class performances of different teams were almost directly proportional to the number of samples available in the training set for each class.  This shows that even though all teams did less well on knot-tying and needle-driving as compared to dissection, additional data for these two tasks can potentially lead to much better performances.

Beyond additional data, other modeling approaches could potentially improve results as well. While all submissions in this challenge leveraged a variant of a 3D CNN architecture, other recent work has also shown that 2D+1 architectures -- 2D CNN layers followed by 1D CNN layers -- can perform well on activity recognition problems \cite{ghadiyaram2019,wu2018}. Moreover, the frame-level pre-processing methods used by team ECBA and SK to segment or emphasize tools is similar to other work which shows that tool detection can aid in automated surgical phase recognition \cite{zisimopoulos2018,sahu2016}. Additional video-level pre-processing, like optical flow calculation, could potentially improve model performances \cite{ullah2018,kumar2016,ladjailia2020,tanberk2020}. 

\iffalse
\textcolor{red}{DISCUSS RECENT WORK HERE THAT MAY BE USED TO SOLVE THIS PROBLEM AND EXTENDS ON METHODS USED BY PARTICIPANTS}
Try 2D+1 method
Attention is all you need
Video action transformer network, Video transformer network 
Optical flow papers?
\fi
\section{Conclusion}

In this paper, we provided results from a challenge that addressed the problem of visual domain adaptation in surgical data science. The models submitted by challenge participants showed some promise in recognizing basic surgical tasks in clinical-like settings when trained on VR surgical data. We found that supplementing the VR training set with even a very small proportion ($\sim$10\%) of clinical-like data dramatically improves the generalizability of the features learned by the models. We expect one of the constraints on the performance of the submitted models was the size of the dataset. However, we believe that VR data and synthetic data is uniquely poised to address this issue in the long-term since the data can be more easily generated and shared than clinical videos. While the problem of surgical visual domain adaptation remains unresolved, we hope that the current work and data encourage further interest and involvement in the problem of visual domain transfer in surgical data science.

\bibliographystyle{unsrt}
\bibliography{references}

\newpage

\appendix
\section{Appendix}
The final design document for the challenge can be found below.


\section*{Supplementary material}

# TASK: SurgVisDom - Surgical visual domain adaptation: from virtual reality to real, clinical environments

## SUMMARY

### Abstract

Provide a summary of the challenge purpose. This should include a general introduction in the topic from both a biomedical as well as from a technical point of view and clearly state the envisioned technical and/or biomedical impact of the challenge.

Surgical data science is revolutionizing minimally invasive surgery. By developing algorithms to be context-aware, exciting applications to augment surgeons are becoming possible. However, there exist many sensitivities around surgical data (or health data more generally) needed to develop context-aware models. This challenge seeks to explore the potential for visual domain adaptation in surgery to overcome data privacy concerns. In particular, we propose to use video from virtual reality simulation data from clinical-like tasks to develop algorithms to recognize activities and then to test these algorithms on videos of the same task in a clinical setting (i.e., porcine model).

### Keywords

List the primary keywords that characterize the task.

visual domain adaptation; surgery; virtual reality; endoscope video; activity recognition

## ORGANIZATION

### Organizers

a) Provide information on the organizing team (names and affiliations).

Aneeq Zia (Intuitive Surgical),
Kiran Bhattacharyya (Intuitive Surgical),
Xi Liu (Intuitive Surgical),
Ziheng Wang (Intuitive Surgical),
Anthony Jarc (Intuitive Surgical)

b) Provide information on the primary contact person.

Aneeq Zia (aneeq.zia@intusurg.com)

### Life cycle type

Define the intended submission cycle of the challenge. Include information on whether/how the challenge will be continued after the challenge has taken place.

Examples:

- One-time event with fixed submission deadline
- Open call
- Repeated event with annual fixed submission deadline

One Time Event

## Challenge venue and platform

a) Report the event (e.g. conference) that is associated with the challenge (if any).

MICCAI

b) Report the platform (e.g. grand-challenge.org) used to run the challenge.

grand-challenge.org

c) Provide the URL for the challenge website (if any).

Part of https://endovis.grand-challenge.org/,  Sub-challenge: https://survisdom.grand-challenge.org/

## Participation policies

a) Define the allowed user interaction of the algorithms assessed (e.g. only (semi-) automatic methods allowed).

Fully Automatic

b) Define the policy on the usage of training data. The data used to train algorithms may, for example, be restricted to the data provided by the challenge or to publicly available data including (open) pre-trained nets.

Open to publicly available data including pre-trained nets. Any privately prepared annotations on public data will need to be released at time of submission.

c) Define the participation policy for members of the organizers' institutes. For example, members of the organizers' institutes may participate in the challenge but are not eligible for awards.

May Not Participate

d) Define the award policy. In particular, provide details with respect to challenge prizes.

3 monetary prizes for 1st, 2nd, and 3rd place. $500, $300 and $200, for 1st,2nd and 3rd, respectively.

e) Define the policy for result announcement.

Examples:
- Top 3 performing methods will be announced publicly.
- Participating teams can choose whether the performance results will be made public.

Top three performing methods will be announced publicly and posted on the website

f) Define the publication policy. In particular, provide details on …
- … who of the participating teams/the participating teams' members qualifies as author
- … whether the participating teams may publish their own results separately, and (if so)
- … whether an embargo time is defined (so that challenge organizers can publish a challenge paper first).

The organizers will publish a challenge paper within six months after the challenge. Following which, the participating teams can publish their own results from the challenge citing the challenge paper. Possibility of a combined publication amongst the participating teams/organization team will also be discussed after the challenge.

## Submission method

a) Describe the method used for result submission. Preferably, provide a link to the submission instructions.

Examples:

- Docker container on the Synapse platform. Link to submission instructions: <URL>
- Algorithm output was sent to organizers via e-mail. Submission instructions were sent by e-mail.

Submission instructions will be posted to the website and sent via email. Results will be submitted via a docker container. The docker containers will need to be sent directly to the organizers with instructions on how to run.

b) Provide information on the possibility for participating teams to evaluate their algorithms before submitting final results. For example, many challenges allow submission of multiple results, and only the last run is officially counted to compute challenge results.

The participants will not be allowed to evaluate their algorithms before submission - only one final submission per team.

## Challenge schedule

Provide a timetable for the challenge. Preferably, this should include

- the release date(s) of the training cases (if any)
- the registration date/period
- the release date(s) of the test cases and validation cases (if any)
- the submission date(s)
- associated workshop days (if any)
- the release date(s) of the results

Release of training cases in June 2020; registration ongoing; submission date September 2020; MICCAI 2020; release of results at MICCAI 2020; release of a subset of test set after MICCAI 2020

## Ethics approval

Indicate whether ethics approval is necessary for the data. If yes, provide details on the ethics approval, preferably institutional review board, location, date and number of the ethics approval (if applicable). Add the URL or a reference to the document of the ethics approval (if available).

An existing Western IRB will be used.

## Data usage agreement

Clarify how the data can be used and distributed by the teams that participate in the challenge and by others during and after the challenge. This should include the explicit listing of the license applied.

Examples:

- CC BY (Attribution)
- CC BY-SA (Attribution-ShareAlike)
- CC BY-ND (Attribution-NoDerivs)
- CC BY-NC (Attribution-NonCommercial)
- CC BY-NC-SA (Attribution-NonCommercial-ShareAlike)
- CC BY-NC-ND (Attribution-NonCommercial-NoDerivs)

CC_BY_NC_ND

Additional comments: (Attribution-NonCommercial-NoDerivs)

## Code availability

a) Provide information on the accessibility of the organizers' evaluation software (e.g. code to produce rankings). Preferably, provide a link to the code and add information on the supported platforms.

open source on challenge site

b) In an analogous manner, provide information on the accessibility of the participating teams' code.

open and private code submission will be accepted

## Conflicts of interest

Provide information related to conflicts of interest. In particular provide information related to sponsoring/funding of the challenge. Also, state explicitly who had/will have access to the test case labels and when.

Sposorship/funding will be done by Intuitive Surgical. The organizers who are affiliated with Intuitive will perform testing.

# MISSION OF THE CHALLENGE

## Field(s) of application

State the main field(s) of application that the participating algorithms target.

Examples:

- Diagnosis
- Education
- Intervention assistance
- Intervention follow-up
- Intervention planning
- Prognosis
- Research
- Screening
- Training
- Cross-phase

Assistance, Research, Training

## Task category(ies)

State the task category(ies).

Examples:

- Classification
- Detection
- Localization
- Modeling
- Prediction
- Reconstruction
- Registration
- Retrieval
- Segmentation
- Tracking

Classification

## Cohorts

We distinguish between the target cohort and the challenge cohort. For example, a challenge could be designed around the task of medical instrument tracking in robotic kidney surgery. While the challenge could be based on ex vivo data obtained from a laparoscopic training environment with porcine organs (challenge cohort), the final biomedical application (i.e. robotic kidney surgery) would be targeted on real patients with certain characteristics defined by inclusion criteria such as restrictions regarding gender or age (target cohort).

a) Describe the target cohort, i.e. the subjects/objects from whom/which the data would be acquired in the final biomedical application.

Robotic clinical-like activities on a porcine model used during training courses; subjects will be surgeons of varying experiences

b) Describe the challenge cohort, i.e. the subject(s)/object(s) from whom/which the challenge data was acquired.

Virtual reality tasks of the same clinical-like activities as the target cohort; subjects will be surgeons and non-surgeons (but with extensive experience operating the robot)

## Imaging modality(ies)

Specify the imaging technique(s) applied in the challenge.

one channel of endscopic video

## Context information

Provide additional information given along with the images. The information may correspond …

a) … directly to the image data (e.g. tumor volume).

Each video clip will include a label of the activity being performed

b) … to the patient in general (e.g. gender, medical history).

not applicable in VR or porcine

## Target entity(ies)

a) Describe the data origin, i.e. the region(s)/part(s) of subject(s)/object(s) from whom/which the image data would be acquired in the final biomedical application (e.g. brain shown in computed tomography (CT) data, abdomen shown in laparoscopic video data, operating room shown in video data, thorax shown in fluoroscopy video). If necessary, differentiate between target and challenge cohort.

The data will be acquired from VR and training courses on porcine models. The clinical-like activity is performed in a standardized manner and is also represented in the VR simulator.

b) Describe the algorithm target, i.e. the structure(s)/subject(s)/object(s)/component(s) that the participating algorithms have been designed to focus on (e.g. tumor in the brain, tip of a medical instrument, nurse in an operating theater, catheter in a fluoroscopy scan). If necessary, differentiate between target and challenge cohort.

Surgical activity recognition during clinical-like task. There will be 3 activities to recognize.

## Assessment aim(s)

Identify the property(ies) of the algorithms to be optimized to perform well in the challenge. If multiple properties are assessed, prioritize them (if appropriate). The properties should then be reflected in the metrics applied (see below, parameter metric(s)), and the priorities should be reflected in the ranking when combining multiple metrics that assess different properties.

- Example 1: Find highly accurate liver segmentation algorithm for CT images.
- Example 2: Find lung tumor detection algorithm with high sensitivity and specificity for mammography images.

Corresponding metrics are listed below (parameter metric(s).

Sensitivity, Accuracy, Precision

Additional points: The assessment aim will be multi-class classification measured by the average f1-score across different classes.

## DATA SETS

## Data source(s)

a) Specify the device(s) used to acquire the challenge data. This includes details on the device(s) used to acquire the imaging data (e.g. manufacturer) as well as information on additional devices used for performance assessment (e.g. tracking system used in a surgical setting).

The Intuitive Data Recorder (IDR) will be used to capture video at 720p and 60fps from one channel of the endoscope on da Vinci surgical systems.

b) Describe relevant details on the imaging process/data acquisition for each acquisition device (e.g. image acquisition protocol(s)).

N/A

c) Specify the center(s)/institute(s) in which the data was acquired and/or the data providing platform/source (e.g. previous challenge). If this information is not provided (e.g. for anonymization reasons), specify why.

Data will be collected at Intuitive Surgical training labs.

d) Describe relevant characteristics (e.g. level of expertise) of the subjects (e.g. surgeon)/objects (e.g. robot) involved in the data acquisition process (if any).

Experience ranges from 0 robotic cases to over 1000 robotic cases.

### Training and test case characteristics

a) State what is meant by one case in this challenge. A case encompasses all data that is processed to produce one result that is compared to the corresponding reference result (i.e. the desired algorithm output).

Examples:

- Training and test cases both represent a CT image of a human brain. Training cases have a weak annotation (tumor present or not and tumor volume (if any)) while the test cases are annotated with the tumor contour (if any).
- A case refers to all information that is available for one particular patient in a specific study. This information always includes the image information as specified in data source(s) (see above) and may include context information (see above). Both training and test cases are annotated with survival (binary) 5 years after (first) image was taken.

Our dataset will consists of 3 commonly performed surgical tasks (needle-driving, knot tying and dissection) in VR and porcine. Training cases will represent individual video clips from VR based skills tasks performed on the da Vinci SimNow platform along with analogous surgical tasks performed on a porcine model. Test cases will include the same surgical tasks performed on a porcine model. Each training video clip will have a single task label, whereas the video clips in the test set can contain more than one surgical task (non-overlapping). The participants will need to produce a frame-by-frame prediction for each video clip in the test set.

b) State the total number of training, validation and test cases.

22 (VR) + 3 (porcine) training cases will be made available for the 3 tasks resulting in a total set of 75 training videos. The test set will contain a total of 30 video clips with a balance representation of the 3 surgical tasks.

c) Explain why a total number of cases and the specific proportion of training, validation and test cases was chosen.

The numbers indicated were kept keeping in mind data collection technicalities and to provide enough data to the participants for developing meaningful models.

d) Mention further important characteristics of the training, validation and test cases (e.g. class distribution in classification tasks chosen according to real-world distribution vs. equal class distribution) and justify the choice.

We will try to ensure class balance within training and testing sets across VR and porcine models. The 3 chosen tasks are equally important and prevalent in surgical training.

### Annotation characteristics

a) Describe the method for determining the reference annotation, i.e. the desired algorithm output. Provide the information separately for the training, validation and test cases if necessary. Possible methods include manual image annotation, in silico ground truth generation and annotation by automatic methods.

If human annotation was involved, state the number of annotators.

1-2 human annotators will be used to temporally annotate porcine and VR video data where required.

b) Provide the instructions given to the annotators (if any) prior to the annotation. This may include description of a training phase with the software. Provide the information separately for the training, validation and test cases if necessary. Preferably, provide a link to the annotation protocol.

Annotators will only specify start and stop times of individual tasks for porcine data - no other special instruction would be given.

c) Provide details on the subject(s)/algorithm(s) that annotated the cases (e.g. information on level of expertise such as number of years of professional experience, medically-trained or not). Provide the information separately for the training, validation and test cases if necessary.

Annotators will be humans who are well-versed with clinical knowledge required for such annotations.

d) Describe the method(s) used to merge multiple annotations for one case (if any). Provide the information separately for the training, validation and test cases if necessary.

One case will be annotated by only one annotator. However, each annotation will be validated and adjusted by an expert as needed.

## Data pre-processing method(s)

Describe the method(s) used for pre-processing the raw training data before it is provided to the participating teams. Provide the information separately for the training, validation and test cases if necessary.

Raw videos will be provided to the participants.

## Sources of error

a) Describe the most relevant possible error sources related to the image annotation. If possible, estimate the magnitude (range) of these errors, using inter-and intra-annotator variability, for example. Provide the information separately for the training, validation and test cases, if necessary.

The surgical tasks chose are relatively simple to annotate and the annotations will be validated by experts. Therefore, we expect the range of annotation error to be very low.

b) In an analogous manner, describe and quantify other relevant sources of error.

To the best of our knowledge, we don't expect any other sources of error.

# ASSESSMENT METHODS

## Metric(s)

a) Define the metric(s) to assess a property of an algorithm. These metrics should reflect the desired algorithm properties described in assessment aim(s) (see above). State which metric(s) were used to compute the ranking(s) (if any).

- Example 1: Dice Similarity Coefficient (DSC)
- Example 2: Area under curve (AUC)

Frame level average F1-score across the 3 classes will be used for assessment.

b) Justify why the metric(s) was/were chosen, preferably with reference to the biomedical application.

Many surgical/non-surgical activity recognition work in the recent few years have shown the validity of this metric.

## Ranking method(s)

a) Describe the method used to compute a performance rank for all submitted algorithms based on the generated metric results on the test cases. Typically the text will describe how results obtained per case and metric are aggregated to arrive at a final score/ranking.

The performance rank will be based on the rank of the f1-score.

b) Describe the method(s) used to manage submissions with missing results on test cases.

The submission will be docker based with model and prediction script requirement from the participants. This will eliminate the possibility of missing results since the organizing team will evaluate performances on test cases.

c) Justify why the described ranking scheme(s) was/were used.

Using f1-score for ranking seems most reasonable for such a classification problem.

## Statistical analyses

a) Provide details for the statistical methods used in the scope of the challenge analysis. This may include

- description of the missing data handling,
- details about the assessment of variability of rankings,
- description of any method used to assess whether the data met the assumptions, required for the particular statistical approach, or
- indication of any software product that was used for all data analysis methods.

The main differentiator of algorithms will be a classification f1-score . However we will also perform statistical significance checks on model performance across tasks and cases.

b) Justify why the described statistical method(s) was/were used.

The ranking will be classification based hence no special statistical analyses method explored.

## Further analyses

Present further analyses to be performed (if applicable), e.g. related to

- combining algorithms via ensembling,
- inter-algorithm variability,
- common problems/biases of the submitted methods, or
- ranking variability.

N/A

## ADDITIONAL POINTS

### References

Please include any reference important for the challenge design, for example publications on the data, the annotation process or the chosen metrics as well as DOIs referring to data or code.

## Further comments

Further comments from the organizers.



\end{document}